\documentclass[final]{cvpr}

\usepackage{times}
\usepackage{epsfig}
\usepackage{graphicx}
\usepackage{amsmath}
\usepackage{amssymb}

\usepackage{color}
\usepackage{cite}
\usepackage{appendix}
\usepackage{pdfpages}
\usepackage{float}

\usepackage{subfigure}
\usepackage[pagebackref=true,breaklinks=true,colorlinks,bookmarks=false]{hyperref}

\def\cvprPaperID{1760} % *** Enter the CVPR Paper ID here
\def\confYear{CVPR 2021}

\begin{document}

%%%%%%%%% TITLE
\title{ClassSR: A General Framework to Accelerate Super-Resolution Networks by Data Characteristic}

\author{Xiangtao Kong$^{1,2}$ \quad\quad\quad Hengyuan Zhao$^{1}$ \quad\quad\quad Yu Qiao$^{1,3}$ \quad\quad\quad Chao Dong$^{1,4}$ \thanks{Corresponding author (e-mail: chao.dong@siat.ac.cn)}\\
$^{1}$Key Laboratory of Human-Machine Intelligence-Synergy Systems,\\
Shenzhen Institutes of Advanced Technology, Chinese Academy of Sciences\\
$^{2}$University of Chinese Academy of Sciences\\
$^{3}$Shanghai AI Lab, Shanghai, China\\
$^{4}$SIAT Branch, Shenzhen Institute of Artificial Intelligence and Robotics for Society\\
{\tt\small \{xt.kong, hy.zhao1, yu.qiao, chao.dong\}@siat.ac.cn}
% For a paper whose authors are all at the same institution,
% omit the following lines up until the closing ``}''.
% Additional authors and addresses can be added with ``\and'',
% just like the second author.
% To save space, use either the email address or home page, not both
}

\maketitle

%%%%%%%%% ABSTRACT
\begin{abstract}
We aim at accelerating super-resolution (SR) networks on large images (2K-8K). The large images are usually decomposed into small sub-images in practical usages. Based on this processing, we found that different image regions have different restoration difficulties and can be processed by networks with different capacities. Intuitively, smooth areas are easier to super-solve than complex textures. To utilize this property, we can adopt appropriate SR networks to process different sub-images after the decomposition. On this basis, we propose a new solution pipeline -- ClassSR that combines classification and SR in a unified framework. In particular, it first uses a Class-Module to classify the sub-images into different classes according to restoration difficulties, then applies an SR-Module to perform SR for different classes. The Class-Module is a conventional classification network, while the SR-Module is a network container that consists of the to-be-accelerated SR network and its simplified versions. We further introduce a new classification method with two losses -- Class-Loss and Average-Loss to produce the classification results. After joint training, a majority of sub-images will pass through smaller networks, thus the computational cost can be significantly reduced. Experiments show that our ClassSR can help most existing methods (e.g., FSRCNN, CARN, SRResNet, RCAN) save up to 50\% FLOPs on DIV8K datasets. This general framework can also be applied in other low-level vision tasks.

\end{abstract}

%%%%%%%%% BODY TEXT
\section{Introduction}
\label{sec:intro}
Image super-resolution (SR) is a long-studied topic, which aims to generate a high-resolution visual-pleasing image from a low-resolution input. In this paper, we study how to accelerate SR algorithms on ``large'' input images, which will be upsampled to at least 2K resolution ($2048 \times 1080$). While in real-world usages, the image/video resolution for smartphones and TV monitors has already reached 4K ($4096 \times 2160$), or even 8K ($7680 \times 4320$). As most recent SR algorithms are built on CNNs, the memory and computational cost will grow quadratically with the input size. Thus it is necessary to decompose input into sub-images and continuously accelerate SR algorithms to meet the requirement of real-time implementation on real images.

\begin{figure}[t!]
  \centering
  \includegraphics[width=0.75\linewidth]{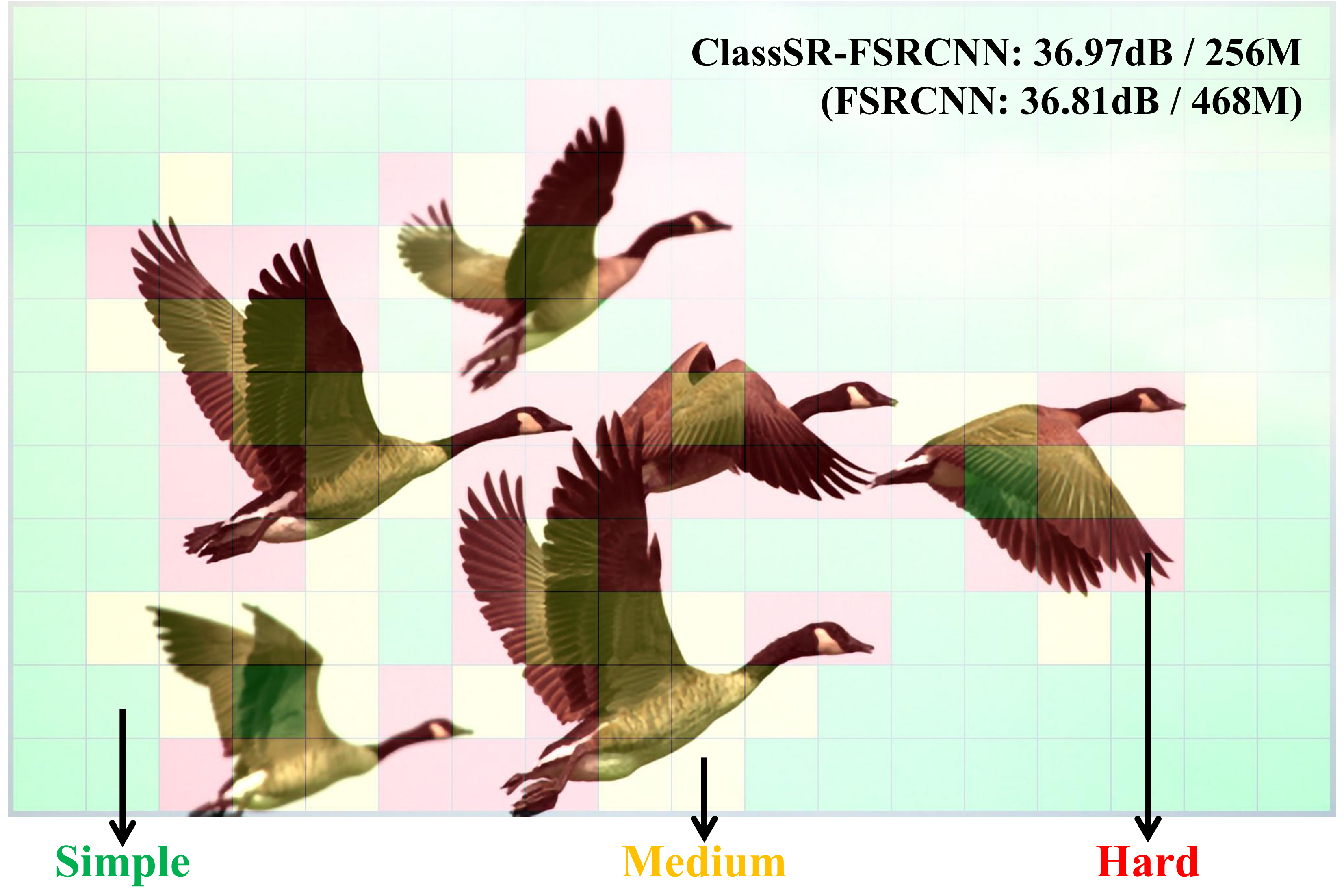} 
  \vskip -0.1cm
  \caption{The SR result (x4) of ClassSR-FSRCNN. The Class-Module classifies the image ``0896'' (DIV2K) into 56\% \textit{simple}, 20\% \textit{medium} and 24\% \textit{hard} sub-images. Compared with FSRCNN, ClassSR-FSRCNN uses only 55\% FLOPs to achieve the same performance.}
  \label{fig:1}
  \vskip -0.7cm
  \end{figure}

Recent works on SR acceleration focus on proposing light-weight network structures, e.g., from the early FSRCNN~\cite{FSRCNN} to the latest CARN~\cite{CARN}, which are detailed in the Sec.~\ref{sec:related}. We tackle this problem from a different perspective. Instead of designing a faster model, we propose a new processing pipeline that could accelerate most SR methods. Above all, we draw the observation that different image regions require different network complexities (see Sec.~\ref{sec:observation}). For example, the flat area (e.g., sky, land) is naturally easier to process than textures (e.g., hair, feathers). This indicates that if we can use smaller networks to treat less complex image regions, the computational cost will be significantly reduced. According to this observation, we can adopt different networks for different contents after decomposition.

Sub-image decomposition is especially beneficial for large images. First, more regions are relatively simple to restore. According to our statistics, about 60\% LR sub-images ($32 \times 32$) belong to smooth regions for DIV8K~\cite{DIV8K} dataset, while the percentage drops to 30\% for DIV2K~\cite{DIV2K} dataset. Thus the acceleration ratio will be higher for large images. Second, sub-image decomposition can help save memory space in real applications, and is essential for low-memory processing chips. It is also plausible to distribute sub-images to parallel processors for further acceleration.

\begin{figure}[t!]
  \centering
  % %\setlength{\abovecaptionskip}{-0cm}
  % \setlength{\belowcaptionskip}{-8pt}
  \includegraphics[width=0.75\linewidth]{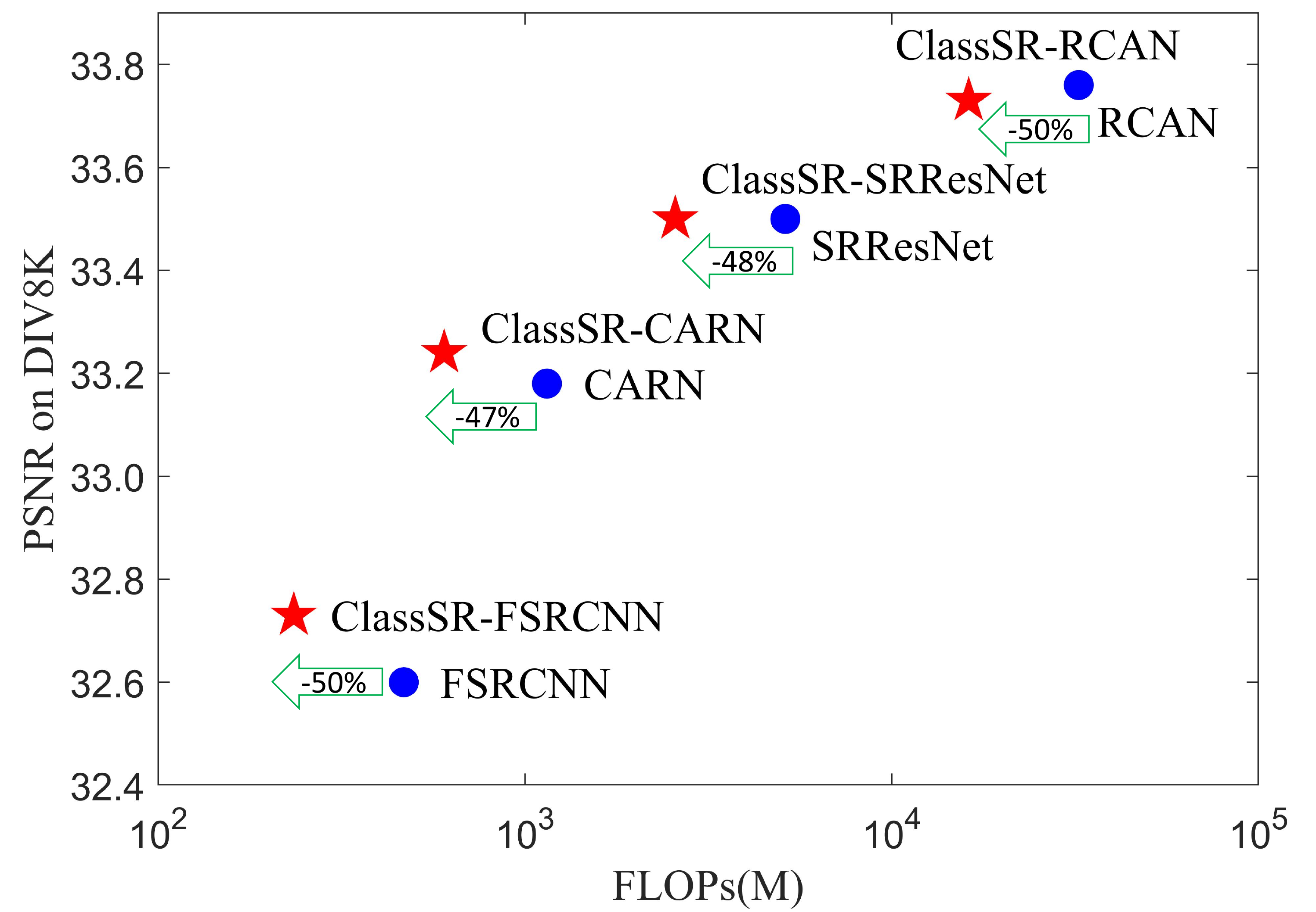} 

  \caption{PSNR and FLOPs comparison between ClassSR and original networks on Test8K with $\times$ 4.}
  \label{fig:2}
  \vskip -0.6cm
\end{figure}

To address the above issue and accelerate existing SR methods, we propose a new solution pipeline, namely ClassSR, to perform classification and super-resolution simultaneously. The framework consists of two modules -- Class-Module and SR-Module. The Class-Module is a simple classification network that classifies the input into a specific class according to the restoration difficulty, while the SR-Module is a network container that processes the classified input with the SR network of the corresponding class. They are connected together and need to be trained jointly. The novelty lies in the classification method and training strategy. Specifically, we introduce two new losses to constrain the classification results. The first one is a Class-Loss that encourages a higher probability of the selected class for individual sub-images. The other one is an Average-Loss that ensures the overall classification results not bias to a single class. These two losses work cooperatively to make the classification meaningful and well-distributed. The Image-Loss ($L_1$ loss) is also added to guarantee the reconstruction performance. For the training strategy, we first pre-train the SR-Module with Image-Loss. Then we fix the SR-Module and optimize the Class-Module with all three losses. Finally, we optimize the two modules simultaneously until convergence. This pipeline is general and effective for different SR networks.

Experiments are conducted on representative SR networks with different scales -- FSRCNN (tiny)~\cite{FSRCNN}, CARN (small)~\cite{CARN}, SRResNet (middle)~\cite{SRResNet} and RCAN (large)~\cite{RCAN}. As shown in Fig.~\ref{fig:2}, the ClassSR method could help these SR networks save 50\%, 47\%, 48\%, 50\% computational cost on the DIV8K dataset, respectively. An example is shown in Fig.~\ref{fig:1}, where the flat areas (color in light green) are processed with the simple network and the textures (color in red) are processed with the complex one. We have also provided a detailed ablation study on the choice of different network settings.

Overall, our contributions are three-fold: (1) \textbf{We propose ClassSR.} It is the first SR pipeline that incorporates classification and super-resolution together on the sub-image level. (2) \textbf{We tackle acceleration by the characteristic of data.} It makes ClassSR orthogonal to other acceleration networks. A network compressed to the limit can still be accelerated by ClassSR. (3) \textbf{We propose a classification method with two novel losses.} It divides sub-images according to their restoration difficulties that are processed by a specific branch instead of predetermined labels, so it can also be directly applied to other low-level vision tasks. The code will be made available: \url{https://github.com/Xiangtaokong/ClassSR}

\section{Related work}
\label{sec:related}

\subsection{CNNs for Image Super-Resolution}
\label{sec:related1}
Since SRCNN~\cite{SRCNN} first introduced convolutional neural networks (CNNs) to the SR task, many deep neural networks have been developed to improve the reconstruction results. For example, VDSR~\cite{VDSR} uses a very deep network to learn the image residual. SRResNet~\cite{SRResNet} introduces ResBlock~\cite{ResNet} to further expand the network size. EDSR~\cite{EDSR} removes some redundant layers from SRResNet and advances results. RDN~\cite{RDN} and RRDB~\cite{ESRGAN} adopt dense connections to utilize the information from preceding layers. Furthermore, RCAN~\cite{RCAN}, SAN~\cite{SAN} and RFA~\cite{RFA} explore the attention mechanism to design deeper networks and constantly refresh the state-of-the-art. However, the expensive computational cost has limited their practical usages.

\subsection{Light-weight SR Networks}
\label{sec:related2}
To reduce computational cost, many acceleration methods have been proposed. FSRCNN~\cite{FSRCNN} and ESPCN~\cite{ESPCN} use the LR image as input and upscale the feature maps at the end of the networks. LapSRN~\cite{LapSRN} introduces a deep laplacian pyramid network that gradually upscales the feature maps. CARN~\cite{CARN} uses the group convolution to design a cascading residual network for fast processing. IMDN~\cite{IMDN} extracts hierarchical features by splitting operations and then aggregates them to save computation. PAN~\cite{PAN} adopts pixel attention to obtain an effective network.

All of those methods aim to design a relatively light-weight network with an acceptable reconstruction performance. In contrast, our ClassSR is a general framework that could accelerate most existing SR methods, even if ranging from tiny networks to large networks.

\subsection{Region-aware Image Restoration}
\label{sec:related3}
Recently, investigators start to treat different image regions with different processing strategies. RAISR~\cite{RAISR} divides the image patches into clusters, and constructs an appropriate filter for each cluster. It also uses an efficient hashing approach to reduce the complexity of the clustering algorithm. SFTGAN~\cite{SFTGAN} introduces a novel spatial feature transform layer to incorporate the high-level semantic prior which is an implicit way to process different regions with different parameters. RL-Restore~\cite{Yu_2018_CVPR} and Path-Restore~\cite{Path-restore} decompose the image into sub-images and estimate an appropriate processing path by reinforcement learning. Different from them, we propose a new classification method to determine the processing of each region.

\section{Methods}
\label{sec:method}

\subsection{Observation}
\label{sec:observation}

\begin{figure}[t!]
  \centering
  % \vskip -0.2cm
  % \setlength{\abovecaptionskip}{-0.0cm}
  % \setlength{\belowcaptionskip}{-0.2cm}
  \includegraphics[width=0.8\linewidth]{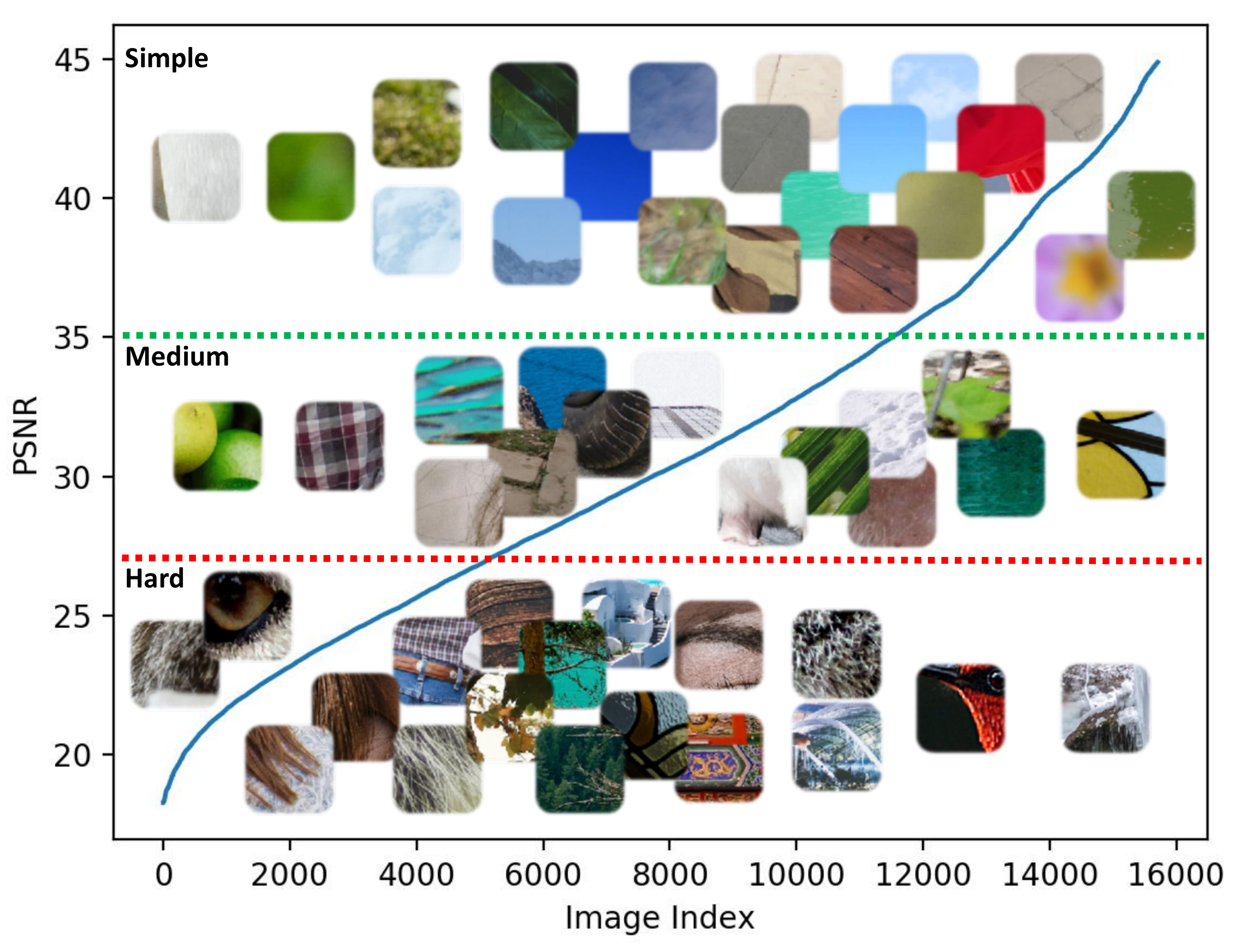} 
  \vskip -0.18cm
     \caption{The ranked PSNR curve of sub-images from DIV2K validation set and the visualization of three classes.}
  \label{fig:3}
  \vskip -0.3cm
\end{figure}

 \begin{table}[!t]
  \small %
  %\centering
  \begin{center}
  \begin{tabular}{|c|cccc|}
  \hline
  Model & FLOPs & Simple & Medium & Hard \\
  \hline\hline
  FSRCNN (16) & 141M  & 42.71dB & -- & --\\
  FSRCNN (36) & 304M & -- & 29.62dB &-- \\
  FSRCNN (56) & 468M & -- & -- & 22.73dB\\
  FSRCNN-O (56) & 468M & 42.70dB & 29.69dB & 22.71dB\\
  \hline
  \end{tabular}
\end{center} 
\vskip -0.2cm
  \caption{PSNR values obtained by three SR branches of ClassSR-FSRCNN with $\times$4. They are separately trained with ``simple, medium, hard'' training data and tested on corresponding validation data. -O: the original networks trained with all data. }
  \label{table:1}
  \vskip -0.5cm
 \end{table}

We first illustrate our observation on different kinds of sub-images. Specifically, we investigate the statistical characteristics of $32 \times 32$ LR sub-images in DIV2K validation dataset~\cite{DIV2K} \footnote{We use 100 validation images (0801-0900), and crop the sub-images with stride 32 and collect 17,808 sub-images in total.}. To evaluate their restoration difficulty, we pass all sub-images through the MSRResNet~\cite{ESRGAN}, and rank these sub-images according to their PSNR values. As depicted in Fig.~\ref{fig:3}, we show these values in a blue curve and separate them into three classes with the same numbers of sub-images -- ``simple, medium, hard''. It is observed that the sub-images with high PSNR values are generally smooth, while the sub-images with low PSNR values contain complex textures.

Then we adopt different networks to deal with different kinds of sub-images. As shown in Table \ref{table:1}, we use three FSRCNN models with the same network structure but different channel numbers in the first conv. layer and the last deconv. layer (i.e., 16, 36, 56). They are separately trained with ``simple, medium, hard'' sub-images from training dataset\footnote{We use 800 training images (0001-0800) in DIV2K, reduce them to 0.6, 0.7, 0.8, 0.9 times, and crop the sub-images with stride 16 and collect 1,594,077 sub-images in total.}. From Table \ref{table:1}, we can find that there is almost no difference for FSRCNN(16) and FSRCNN-O(56) on ``simple'' sub-images, and FSRCNN(36) can achieve roughly the same performance as FSRCNN-O(56) on ``medium'' sub-images. This indicates that we can use a light-weight network to deal with simple sub-images to save computational cost. That is why we propose the following ClassSR method, which could treat different image regions differently and accelerate existing SR methods.

\subsection{Overview of ClassSR}
\label{sec:overview}

\begin{figure*}[ht!]
  \centering
  % %\setlength{\abovecaptionskip}{-0cm}
  % \setlength{\belowcaptionskip}{-0.3cm}
  \includegraphics[width=0.75\linewidth]{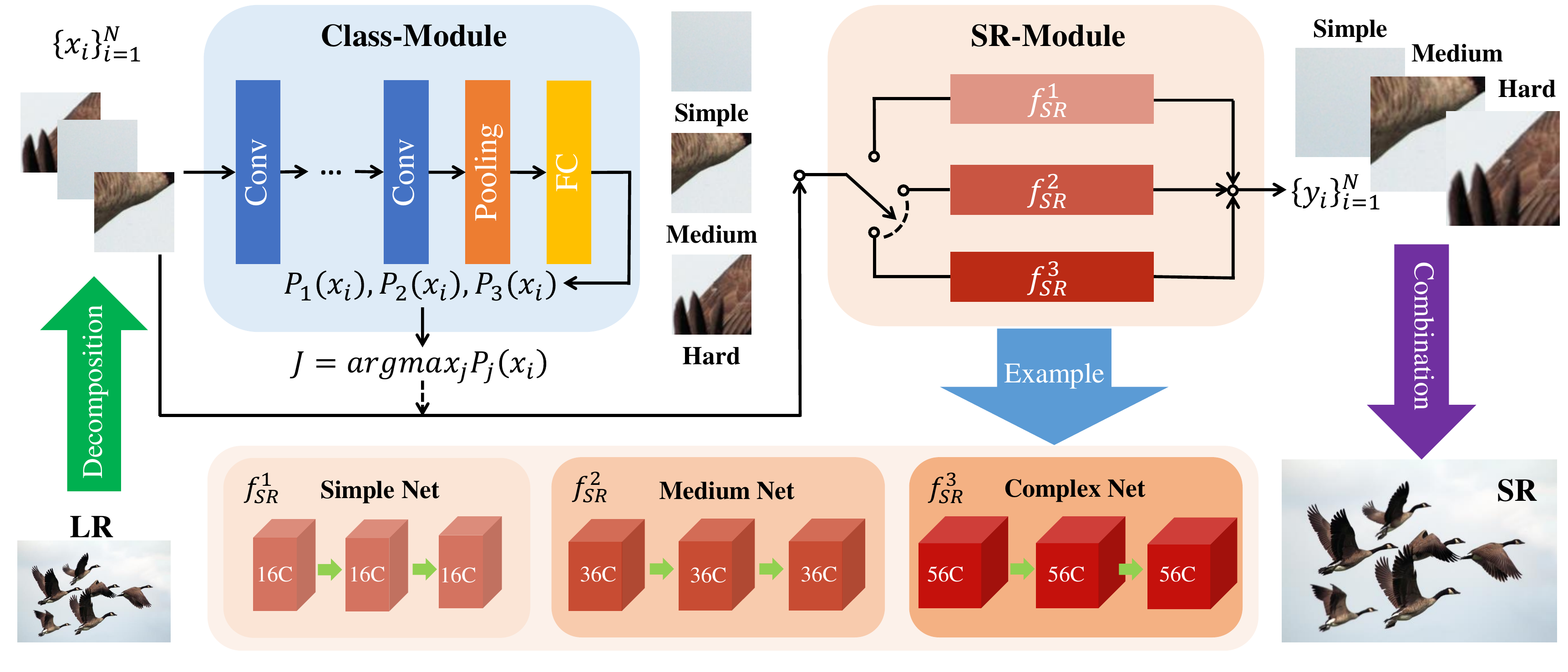} 
  \vskip -0.3cm
    \caption{The overview of the proposed ClassSR, when the number of classes $M=3$. Class-Module: aims to generate the probability vector, SR-Module: aims to deal with the corresponding sub-images.}
  \label{fig:4}
  \vskip -0.5cm
  \end{figure*}

ClassSR is a new solution pipeline for single image SR. It consists of two modules -- Class-Module and SR-Module, as shown in Fig.~\ref{fig:4}. The Class-Module classifies the input images into $M$ classes, while the SR-Module contains $M$ branches (SR networks) $\{f_{SR}^j\}_{j=1}^M$ to deal with different inputs. To be specific, the large input LR image $X$ is first decomposed into overlapping sub-images $\{x_i\}_{i=1}^N$. The Class-Module accepts each sub-image $x_i$ and generates a probability vector $[P_1(x_i),...,P_M(x_i)]$. After that, we determine which SR network to be used by selecting the index of the maximum probability value $J=\arg\max_j P_j(x_i)$. Then $x_i$ will be processed by the $J$th branch of the SR-Module: $y_i=f_{SR}^J(x_i)$. Finally, we combine all output sub-images $\{y_i\}_{i=1}^N$ to get the final large SR image $Y$ (2K-8K).

\subsection{Class-Module}
\label{sec:Class-Module}
The goal of Class-Module is to tell ``whether the input sub-image is easy or hard to reconstruct'' by low-level features. As shown in Fig.~\ref{fig:4}, we design the Class-Module as a simple classification network, which contains five convolution layers, an average pooling layer and a fully-connected layer. The convolution layers are responsible for feature extraction, while the pooling and fully-connected layers output the probability vector. This network is pretty light-weight, and brings little additional computational cost. Experiments show that such a simple structure can already achieve satisfactory classification results. 

\subsection{SR-Module}
\label{sec:SR-Module}
The SR-Module is designed as a container that consists of several independent branches $\{f_{SR}^j\}_{j=1}^M$. In general, each branch can be any learning-based SR network. As our goal is to accelerate an existing SR method (e.g., FSRCNN, CARN), we adopt this SR network as the base network, and set it as the most complex branch $f_{SR}^{M}$. The other branches are obtained by reducing the network complexity of $f_{SR}^{M}$. For simplicity, we use the number of channels in each convolution layer to control the network complexity. Then how many channels are required for each SR branch? The principle is that the branch network should achieve comparable results as the base network trained with all data in the corresponding class. For instance (see Table~\ref{table:1} and Fig.~\ref{fig:4}), the number of channels for $f_{SR}^{1}, f_{SR}^{2}, f_{SR}^{3}$ can be 16, 36, 56, where 56 is the channel number of the base network. Note that we can also decrease the network complexity in other ways, such as reducing layers (see Sec.~\ref{sec:layers}), as long as the network performance meets the above principle.

\subsection{Classification Method}
\label{sec:Training Processing}

During training, the Class-Module classifies sub-images according to their restoration difficulties of a specific branch instead of predetermined labels. Therefore, different from testing, the input sub-image $x$ should pass through all $M$ SR branches. Besides, in order to ensure that the Class-Module can accept the gradient propagation from the reconstruction results, we multiply the reconstructed sub-images $f_{SR}^i(x)$ and the corresponding classification probability $P_i(x)$ to generate the final SR output $y$ as:

\begin{equation}
  \begin{aligned}
    y=\sum_{i=1}^{M} P_i(x) \times f_{SR}^i(x).
  \end{aligned}
  \label{equ:4}
\end{equation}

We just use Image-Loss ($L_1$ loss) to constrain $y$, then we can obtain classification probabilities automatically. But during testing, the input only pass the SR branch with the maximum probability. Thus, we propose $L_{c}$ (Class-Loss, see Sec.~\ref{sec:Class-Loss}) to make the maximum probability to approach 1, and $y$ will be equal to the sub-image with probability 1. Note that if we only adopt the Image-Loss and Class-Loss, the training will easily converge to an extreme point, where all images are classified into the most complex branch. To avoid such a biased result, we design the $L_{a}$ (Average-Loss, see Sec.~\ref{sec:Average-Loss}) to constrain the classification results. This is our proposed new classification method.

\subsection{Loss Functions}
\label{sec:loss}

The loss function consists of three losses -- a commonly used $L_1$ loss (Image-Loss) and our proposed two losses $L_{c}$ (Class-Loss) and $L_{a}$ (Average-Loss). Specifically, $L_1$ is used to ensure the image reconstruction quality, $L_c$ improves the effectiveness of classification, and $L_a$ ensures that each SR branch can be chosen equally. The loss function is shown as:
\begin{equation}
  \begin{aligned}
    L=w_1 \times L_{1} + w_2 \times L_{c} + w_3 \times L_a,
  \end{aligned}
  \label{equ:5}
\end{equation}
where $w_1$, $w_2$ and $w_3$ are the weights to balance different loss terms. $L_1$ is the 1-norm distance between the output image and ground truth, just as in previous works ~\cite{VDSR, SRResNet}. The two new losses $L_{c}$ and $L_{a}$ are detailed below.

\subsubsection{Class-Loss}
\label{sec:Class-Loss}

As mentioned in Sec.~\ref{sec:Training Processing}, the Class-Loss constrains the output probability distribution of the Class-Module. We prefer that the Class-Module has much higher confidence in class with the maximum probability than others. For example, the classification result [0.90, 0.05, 0.05] is better than [0.34,0.33,0.33], as the latter seems like a random selection. The Class-Loss is formulated as:
\begin{equation}
  \begin{aligned}
    L_{c}=-\sum_{i=1}^{M-1}\sum_{j=i+1}^{M}|P_i(x)-P_j(x)|,s.t. \sum_{i=1}^{M}P_i(x)=1.
\end{aligned}
  \label{equ:6}
\end{equation}
where $M$ is the number of classes. The $L_c$ is the negative number of distance sum between each class probability for a same sub-image. This loss can greatly enlarge the probability gap between different classification results so that the maximum probability value will be close to 1. 

\subsubsection{Average-Loss}

\label{sec:Average-Loss}

As mentioned in Sec.~\ref{sec:Training Processing}, if we only adopt the Image-Loss and Class-Loss, the sub-images are prone to be assigned to the most complex branch. This is because that the most complex SR network can easily get better results. Then the Class-Module will lose its functionality and the SR-Module degenerates to the base network. To avoid this, we should ensure that each SR branch has an equal opportunity to be selected. Therefore, we design the Average-Loss to constrain the classification results. It is formulated as:
\begin{equation}
  \begin{aligned}
    L_a=\sum_{i=1}^{M}|\sum_{j=1}^{B}P_i(x_j)-\frac{B}{M}|,
  \end{aligned}
  \label{equ:8}
\end{equation}
where $B$ is the batch size. The $L_a$ is the sum of the distance between the average number ($\frac{B}{M}$) and the sub-images number of each class within a batch. We use the probability sum $\sum_{j=1}^{B}P_i(x_j)$ to calculate the sub-images number because statistic number do not propagate gradients. With this loss, the number of sub-images that pass through each SR branch during training would be approximately the same.

\subsection{Training Strategy}
\label{sec:Training Strategy}

We propose to train the ClassSR by three steps: First, pre-train SR-Module, then train Class-Module with fixing SR-Module using the proposed three losses, finally finetune all networks jointly. This is because that if we train both Class-Module and SR-Module from scratch, the performance will be very unstable, and the classification will easily fall into a bad local minimum.

To pre-train the SR-Module, we use the data classified by the PSNR values. Specifically, all sub-images are passed through a well-trained MSRResNet. Then these sub-images are ranked according to their PSNR values. Next, the first 1/3 sub-images are assigned to the hard class, while the last 1/3 belong to the simple class, just as in Sec.~\ref{sec:observation}. Then we train the simple/medium/complex SR branch on the corresponding simple/medium/hard data. Although using PSNR obtained by MSRResNet to estimate the restoration difficulties is not perfect for different SR branches, it could provide SR branches a good starting point. 

After that, we add the Class-Module and fix the parameters of the SR-Module. The overall model is trained with the three losses on all data. As shown in Fig.~\ref{fig:cla} and Fig.~\ref{fig:clb}, this procedure could give the Class-Module a primary classification ability. 

Afterwards, we relax all parameters and finetune the whole model. During joint training, the Class-Module refines its output probability vectors by the final SR results, and the SR-Module updates according to the new classification results. In experiments (see Fig.~\ref{fig:cl}), we can find that the sub-images are assigned to different SR branches, while the performance and efficiency improve simultaneously. 

\subsection{Discussion}

We further clarify the unique features of ClassSR as follows. 1) The classification+SR strategy adopted by ClassSR has significant practical values. This is based on the observation that large images SR (2K-8K) have different characteristics with small images SR (e.g., the same content cover more pixels), thus are more suitable for sub-image decomposition and special treatment. 2) While the idea of divide-and-conquer is straightforward, the novelty of our method lies in the joint optimization of classification and super-resolution. With a unified framework, we can simultaneously constrain the classification and reconstruction results by a dedicated loss combination. 3) ClassSR can be used together with previous methods for double acceleration.

\section{Experiments}
\label{sec:Experiments}

\subsection{Setting}
\label{sec:Setting}

\subsubsection{Training Data}
\label{sec:Training Data}

We use the DIV2K~\cite{DIV2K} dataset for training. To prepare the training data, we first downsample\footnote{We use bicubic downsampling for all experiments.} the original images with scaling factors 0.6, 0.7, 0.8, 0.9 to generate the HR images. These images are further downsampled 4 times to obtain the LR images. Then we densely crop 1.59M sub-images with size $32\times32$ from LR images. These sub-images are equally divided into three classes (0.53M for each) according to their PSNR values through MSRResNet~\cite{ESRGAN}. All sub-images are further augmented by flipping and rotation. Finally, we obtain ``simple, medium, hard'' datasets for SR-Module pre-training. Besides, we also select ten images (index 0801-0810) from the DIV2K validation set for validation during training. 

\begin{figure*}[ht!]
\begin{center}
\includegraphics[width=0.80\linewidth]{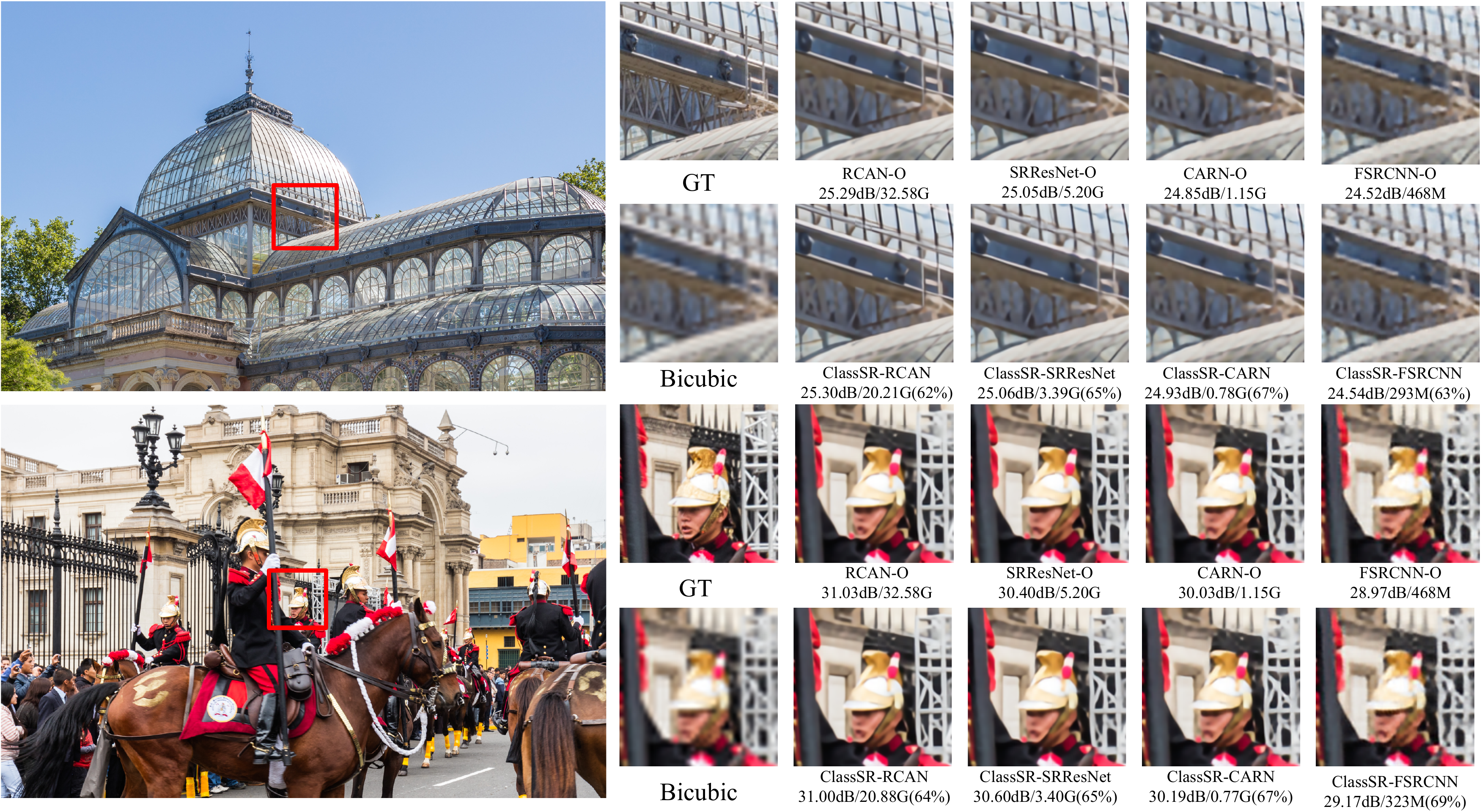} 
\end{center}
\vskip -0.5cm
\caption{Visual results of ClassSR and the original networks on 4K images with $\times$4 super-resolution. The right images are 200$\times$200 which contain decomposition borders.(The size of super-resolved sub-images is 128$\times$128.) -O: the original networks.}
\label{fig:5}
\vskip -0.3cm
\end{figure*}

\begin{table*}[ht!]
  \small
  \begin{center}
  \begin{tabular}{|c|c|cc|cc|cc|}
  \hline
  Model & Parameters& Test2K& FLOPs & Test4K & FLOPs &Test8K & FLOPs  \\
  \hline\hline
  FSRCNN-O  & 25K & 25.61dB & 468M(100\%) & 26.90dB & 468M(100\%) &32.66dB & 468M(100\%) \\
  ClassSR-FSRCNN & 113K & \textcolor{red}{25.61dB} & \textcolor{blue}{311M(66\%)} & \textcolor{red}{26.91dB} & \textcolor{blue}{286M(61\%)} &\textcolor{red}{32.73dB} & \textcolor{blue}{238M(51\%)} \\
  \hline
  CARN-O &295K& 25.95dB & 1.15G(100\%) & 27.34dB & 1.15G(100\%) &33.18dB & 1.15G(100\%) \\
  ClassSR-CARN &645K& \textcolor{red}{26.01dB} & \textcolor{blue}{814M(71\%)} & \textcolor{red}{27.42dB} & \textcolor{blue}{742M(64\%)} &\textcolor{red}{33.24dB} & \textcolor{blue}{608M(53\%)} \\
  \hline
  SRResNet-O &1.5M& 26.19dB & 5.20G(100\%) & 27.65dB & 5.20G(100\%) &33.50dB & 5.20G(100\%)\\
  ClassSR-SRResNet &3.1M& \textcolor{red}{26.20dB} & \textcolor{blue}{3.62G(70\%)} & \textcolor{red}{27.66dB} & \textcolor{blue}{3.30G(63\%)} &\textcolor{red}{33.50dB} & \textcolor{blue}{2.70G(52\%)}\\
  
  \hline  
  RCAN-O & 15.6M & 26.39dB & 32.60G(100\%) & \textcolor{red}{27.89dB} & 32.60G(100\%) &\textcolor{red}{33.76dB} & 32.60G(100\%) \\
  ClassSR-RCAN & 30.1M & \textcolor{red}{26.39dB}  & \textcolor{blue}{21.22G(65\%)} &  27.88dB  & \textcolor{blue}{19.49G(60\%)}  & 33.73dB  & \textcolor{blue}{16.36G(50\%)} \\
  \hline
  \end{tabular}
  \end{center}
  \vskip -0.2cm
  \caption{PSNR values on Test2K, Test4K and Test8K. -O: the original networks. Red/Blue text: best performance/lowest FLOPs.}
  \vskip -0.5cm
  \label{table:2}
  \end{table*}

\subsubsection{Testing Data}

Instead of commonly used SR test sets, such as Set5~\cite{set5} and Set14~\cite{set14}, as their images are too small to be decomposed, we select 300 images (index 1201-1500) from the DIV8K~\cite{DIV8K} dataset. Specifically, the first two hundred images are downsampled to 2K and 4K resolution, respectively, which are used as HR images of Test2K and Test4K datasets. The last hundred images form the Test8K dataset. The LR images are also obtained by $\times4$ downsampling based on HR images. During testing, the LR images are cropped into $32\times32$ sub-images with stride 28. The super-resolved sub-images are combined to SR images by averaging overlapping areas. We use PSNR values between SR and HR images to evaluate the reconstruction performance and calculate the average FLOPs of all $32 \times 32$ sub-images within a test set to evaluate the computational cost.

\subsubsection{Training Details} 

First, we pre-train the SR-Module. The $f_{SR}^{1}$, $f_{SR}^{2}$ and $f_{SR}^{3}$ are separately trained on different training data (``simple, medium, hard''). The mini-batch size is set to 16. $L_1$ loss function~\cite{L1} is adopted with Adam optimizer~\cite{ADAM} ($\beta_1$ = 0.9, $\beta_2$ = 0.999). The cosine annealing learning strategy is applied to adjust the learning rate. The initial learning rate is set to $10^{-3}$ and the minimum is set to $10^{-7}$. The period of cosine is 500k iterations. Then we train the Class-Module with three losses (the weights $w_1, w_2, w_3$ are set to 2000, 1, 6) on all data. Note that we use a larger batch size(96), since the Average-loss needs to balance the number of sub-images within each batch. The other settings are the same as pre-training. The Class-Module is trained within 200k iterations. Finally, we train two modules jointly with all settings unchanged. Besides, we also train the original network with all data in a larger number of iterations than ClassSR for a fair comparison. All models are built on the PyTorch framework~\cite{pytorch} and trained with NVIDIA 2080Ti GPUs.

\subsection{ClassSR with Existing SR networks}
\label{sec:ClassSR with Existing SR networks}

ClassSR is a general framework that can incorporate most deep learning based SR methods, regardless of the network structure. Thus, we do not compare ClassSR with other network accelerating strategies because they can also be further accelerated by ClassSR. Therefore, to demonstrate its effectiveness, we use the ClassSR to accelerate FSRCNN (tiny)~\cite{FSRCNN}, CARN (small)~\cite{CARN}, SRResNet (middle)~\cite{SRResNet} and RCAN (large)~\cite{RCAN}, which are representative networks of different network scales. Their SR-Modules all contain three branches. The most complex branch $f_{SR}^{3}$ is the original network, while the other branches are obtained by reducing the channels in each convolution layer. Specifically, the channel configurations of the three branches are (16, 36, 56) for FSRCNN\footnote{As FSRCNN has different numbers of channels in each layer, we only change the first conv. layer and the last deconv. layer.}, (36, 52, 64) for CARN, (36, 52, 64) for SRResNet, and (36, 50, 64) for RCAN. Training and testing follow the same procedure as described above.

Results are summarized in Table \ref{table:2}. Obviously, most ClassSR methods can obtain better performance than the original networks with lower computational cost, ranging from 70\% to 50\%. The reduction of FLOPs is highly correlated with the image resolution of test data. The acceleration on Test8K is the most significant, nearly 2 times (50\% FLOPs) for all methods. This is because a larger input image can be decomposed into more sub-images, which have a higher probability to be processed by simple branches.

\begin{figure}[ht!]

  \centering
  \subfigure[The PSNR curve of Class.]{
  \label{fig:cla}
  \includegraphics[height=2.5 cm]{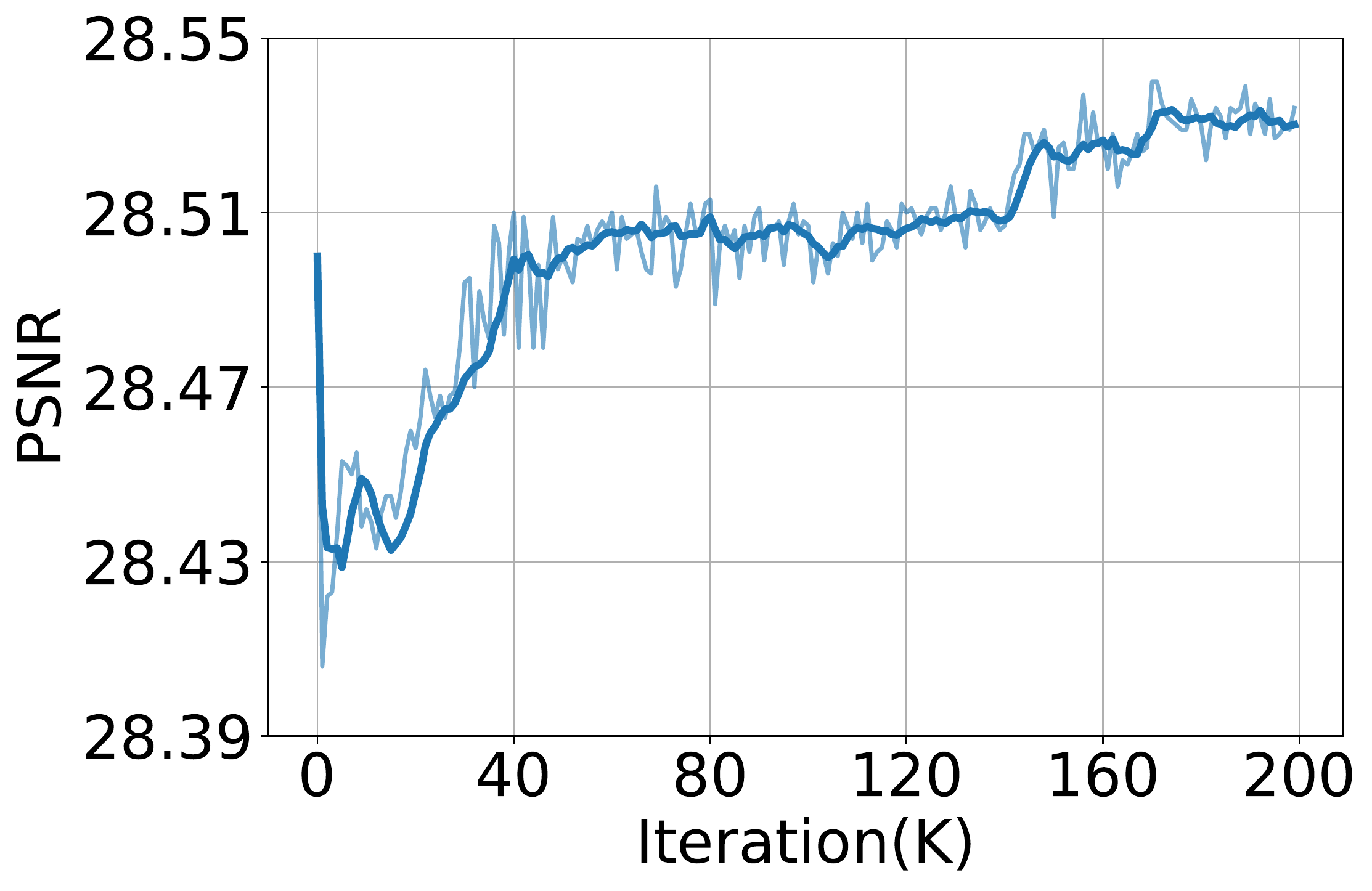}}
  \subfigure[The FLOPs curve of Class.]{
  \label{fig:clb}
  \includegraphics[height=2.5 cm]{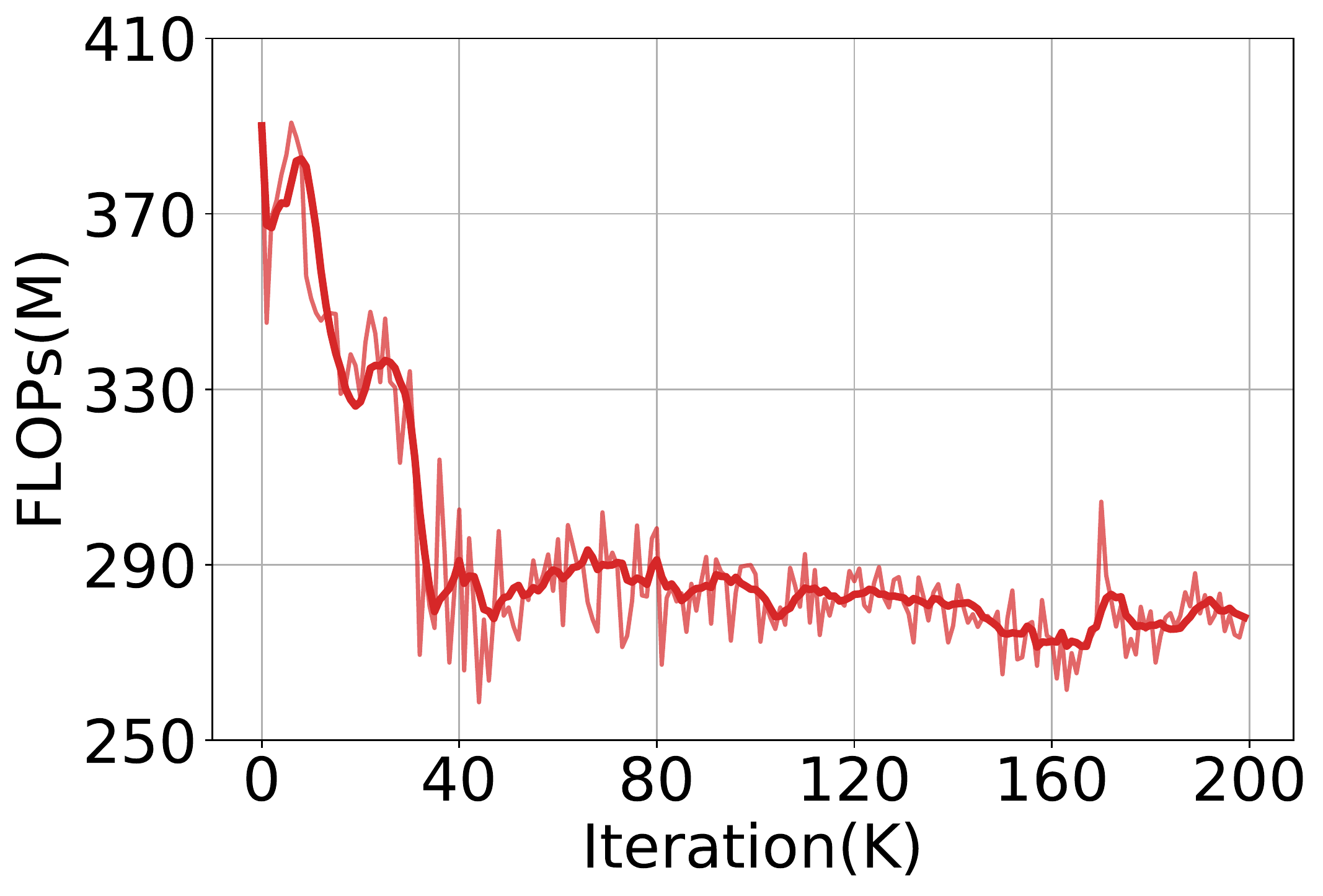}}

  \vskip -0.28cm
  \subfigure[The PSNR curve of Joint.]{
  \label{fig:clc}
  \includegraphics[height=2.5 cm]{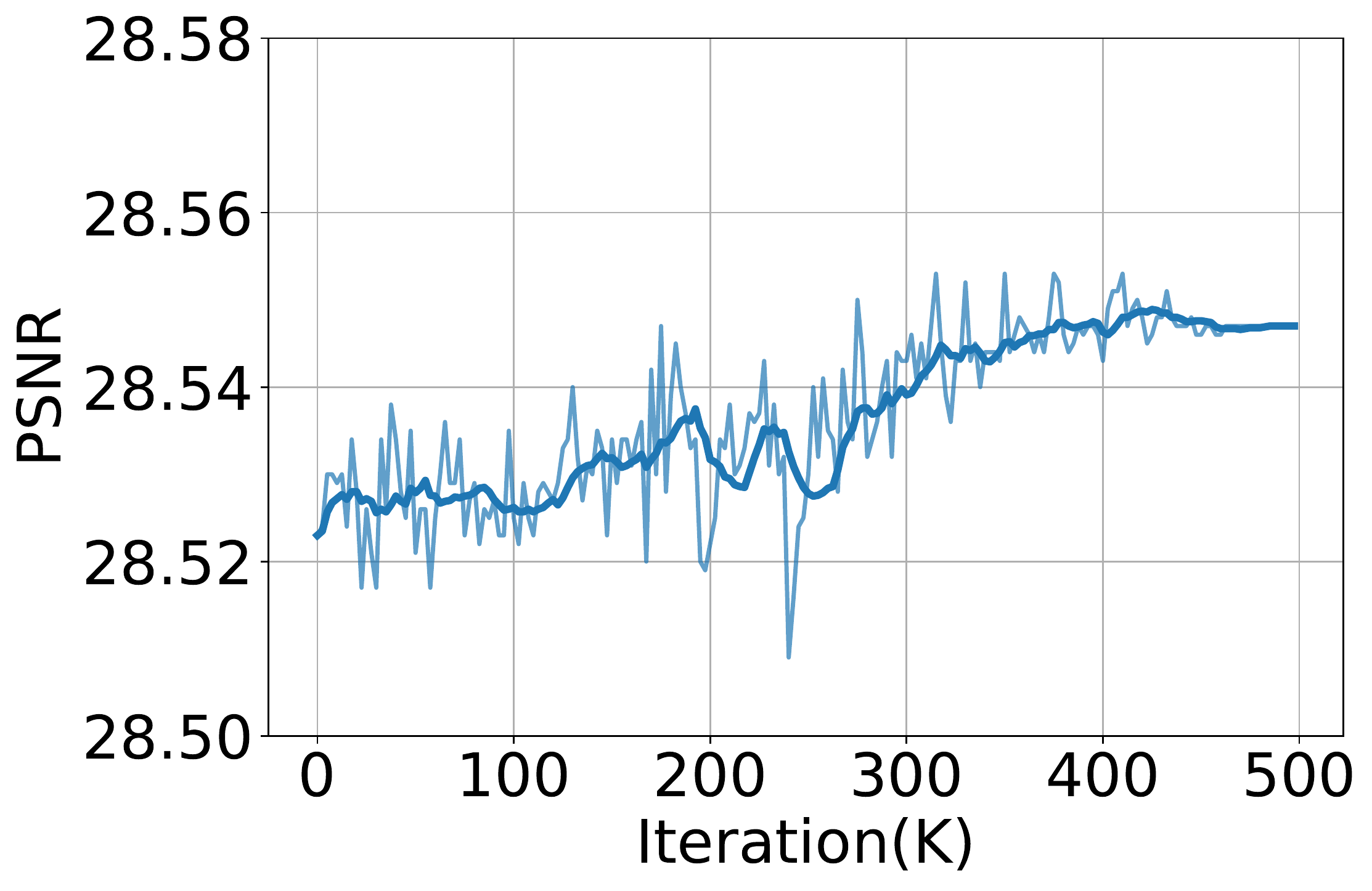}}
  \subfigure[The FLOPs curve of Joint.]{
  \label{fig:cld}
  \includegraphics[height=2.5 cm]{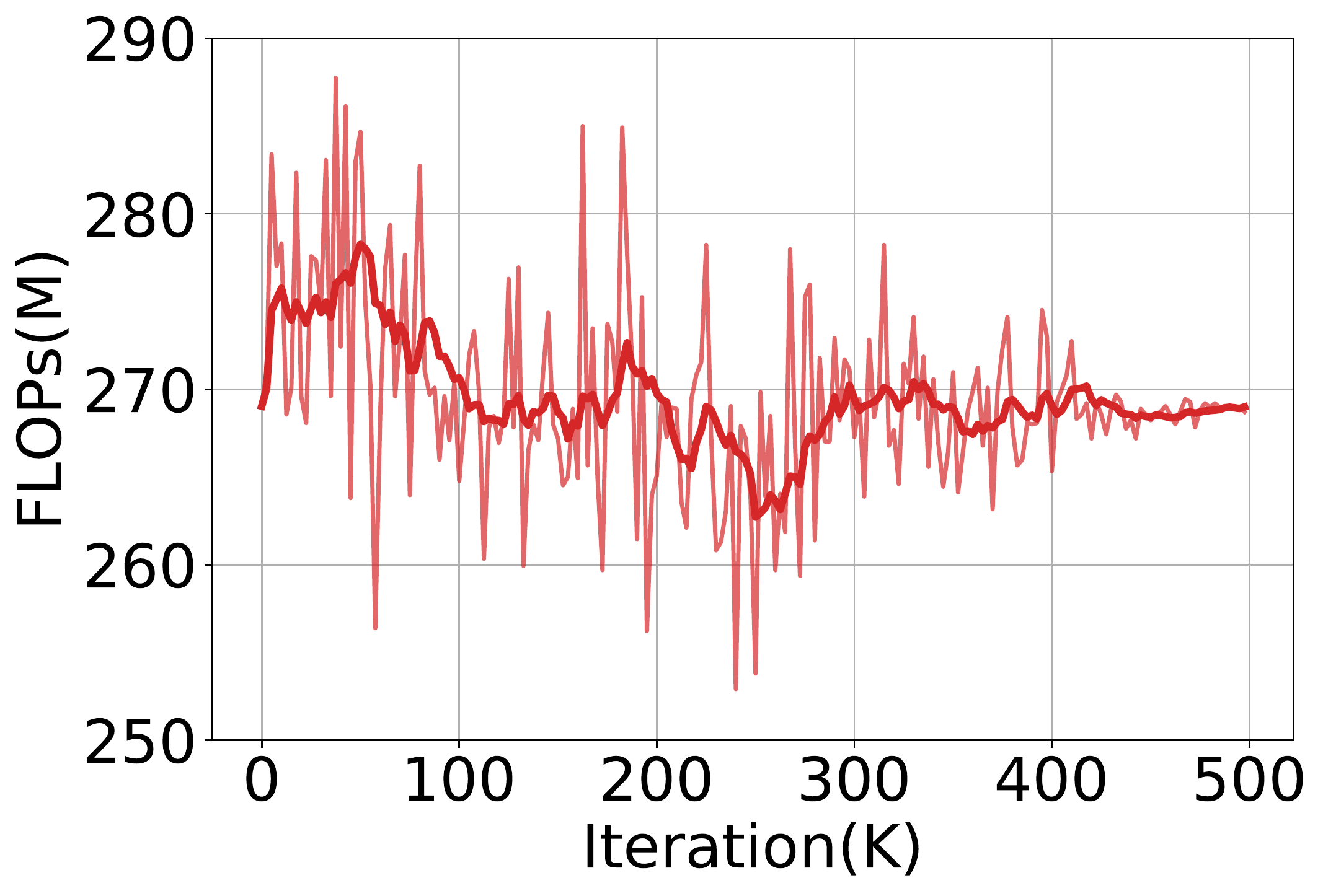}}

  \caption{The training curves of Class-Module (Class) and joint training (Joint) of ClassSR-FSRCNN.}
  \label{fig:cl}
  \vskip -0.3cm
  \end{figure}

To further understand how ClassSR works, we use ClassSR-FSRCNN to illustrate the behaviors and intermediate results of different training stages. First, let us see the performance of SR-Module pre-training. As shown in Table ~\ref{table:1}, the results of SR branches in the corresponding validation sets are roughly the same as the original network. This is in accordance with our observation in Sec.~\ref{sec:observation}. Then we show the validation curves of training Class-Module and joint training in Fig.~\ref{fig:cl}. We can see that the PSNR values increase with the decrease of FLOPs even during the training of Class-Module. This indicates that the increase in performance is not at the cost of the computation burden. In other words, the input images are classified into more appropriate branches during the training process, demonstrating the effectiveness of both two training procedures. After training, we test ClassSR-FSRCNN on Test8K. Statistically, 61\%, 23\%, 16\% sub-images are assigned to FSRCNN (16), FSRCNN (36), FSRCNN (56), respectively. The overall FLOPs drop from 468M to 236M. This further reflects the effectiveness of classification.

Fig.~\ref{fig:5} shows a visual example, where we observe that ClassSR methods can obtain the same visual effects as the original networks. Furthermore, the transitions among different regions are smooth and natural. In other words, treating different regions differently will bring no incoherence between adjacent sub-images.

{\bf Complexity Analysis} 
During testing, first, we use the average FLOPs of all $32 \times 32$ sub-images within a test set to evaluate the running time because the FLOPs is device-independent and well-known by most researchers and engineers. The FLOPs already includes the cost of Class-Module, which is only 8M, almost negligible for the whole model. Second, we need to clarify that the aim of ClassSR is to save FLOPs instead of parameters. The former one can represent the real running time, while the latter one mainly influences the memory. Note that the memory cost brought by model parameters is much less than saving intermediate features, thus the increased parameters brought by ClassSR are acceptable.

  \begin{table*}[ht]
    \small 
      \begin{center}
      \begin{tabular}{|c|cc|cc|cc|}
      \hline
      Model & Test2K& FLOPs & Test4K & FLOPs &Test8K & FLOPs  \\
      \hline\hline
      ClassSR-FSRCNN(2) (16, 56) & 25.61dB & 310M(66\%) & 26.91dB & 280M(60\%) &32.72dB & 228M(49\%)\\
      ClassSR-FSRCNN(3) (16, 36, 56)& 25.61dB & 311M(66\%) & 26.91dB & 286M(61\%) &32.73dB & 238M(51\%)\\
      ClassSR-FSRCNN(4) (16, 29, 43, 56)& 25.61dB & 298M(64\%) & 26.92dB & 290M(62\%) &32.73dB & 238M(51\%)\\
      ClassSR-FSRCNN(5) (16, 26, 36, 46, 56)& 25.63dB & 306M(65\%) & 26.93dB & 286M(61\%) &32.74dB & 248M(53\%)\\
      \hline
      \end{tabular}
      \end{center}
      \vskip -0.25cm
      \caption{PSNR obtained by ClassSR. ClassSR-FSRCNN(2) (16, 56): ClassSR has 2 branches. $f_{SR}^1$ has 16 channels, $f_{SR}^2$ has 56 channels.}
      \label{table:3}
      \vskip -0.25cm
    \end{table*}

    \begin{table*}[ht]
      \small 

        \begin{center}
        \begin{tabular}{|c|cc|cc|cc|}
        \hline
        Model & Test2K& FLOPs & Test4K & FLOPs &Test8K & FLOPs  \\
        \hline\hline
        ClassSR-SRResNet(38 12, 54 14, 64 16) & 26.20dB & 3.60G(69\%) & 27.65dB & 3.28G(63\%) &33.50dB & 2.68G(52\%)\\
        ClassSR-SRResNet(42 8, 56 12, 64 16)& 26.20dB & 3.60G(69\%) & 27.65dB & 3.28G(63\%) &33.50dB & 2.68G(52\%) \\
        \hline
        \end{tabular}
        \end{center}
        \vskip -0.2cm
        \caption{PSNR values obtained by ClassSR with different layers and channels on Test2K, Test4K and Test8K. ClassSR-SRResNet (38 12, 54 14, 64 16): $f_{SR}^1$ has 42 channels and 12 layers, $f_{SR}^2$ has 54 channels and 14 layers, $f_{SR}^3$ has 64 channels and 16 layers.}
        \vskip -0.6cm
        \label{table:5}
        
      \end{table*}

\subsection{Ablation Study}
\label{sec:Ablation Study}

\subsubsection{Effect of Class-Loss}

\label{sec:Effect of Class-Loss}

\begin{figure}[ht!]
  \centering
  \subfigure[PSNR curves]{
  \label{fig:subclassa}
  \includegraphics[height=2.5 cm]{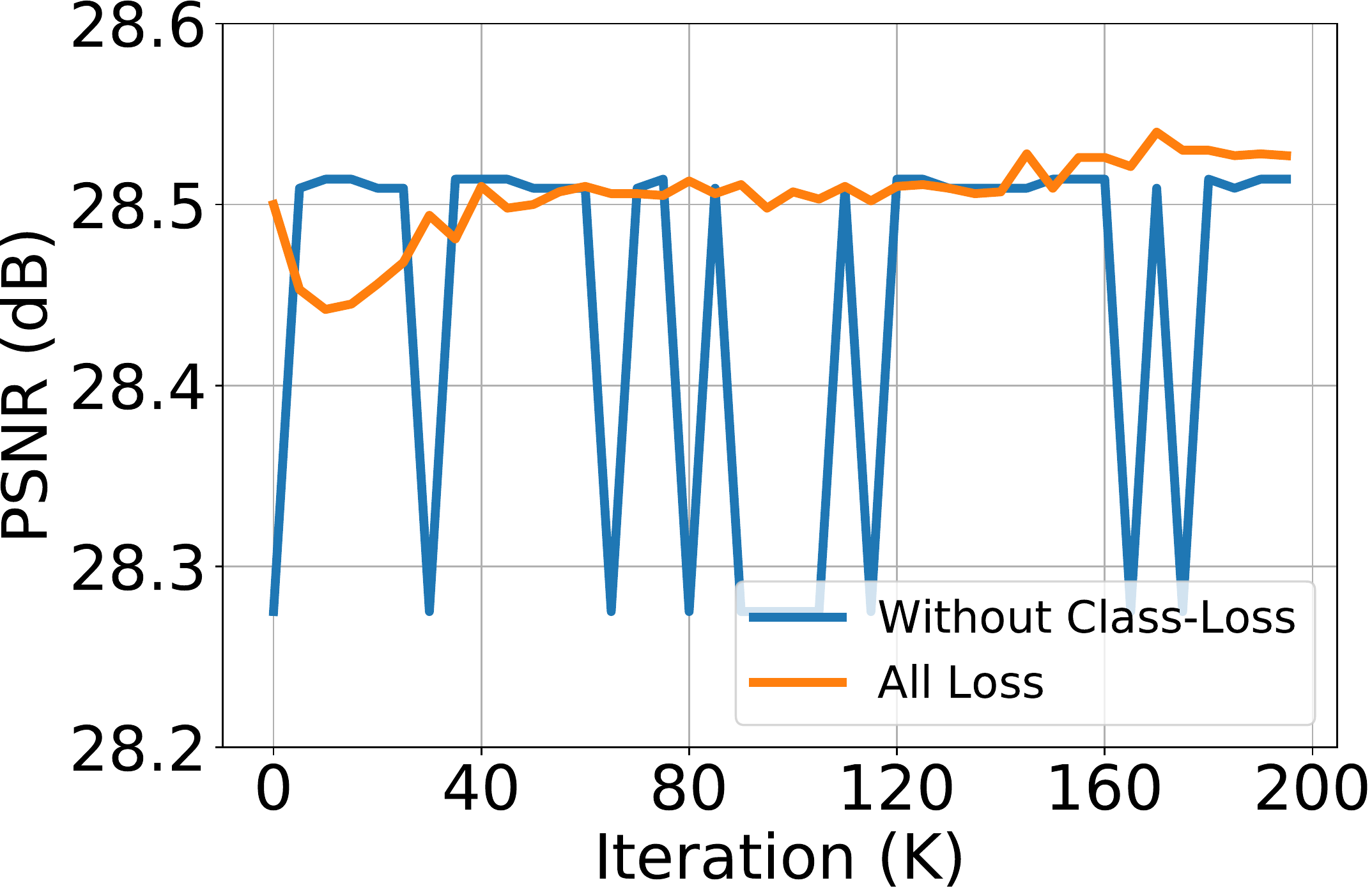}}
  \subfigure[FLOPs curves]{
  \label{fig:subclassb}
  \includegraphics[height=2.5 cm]{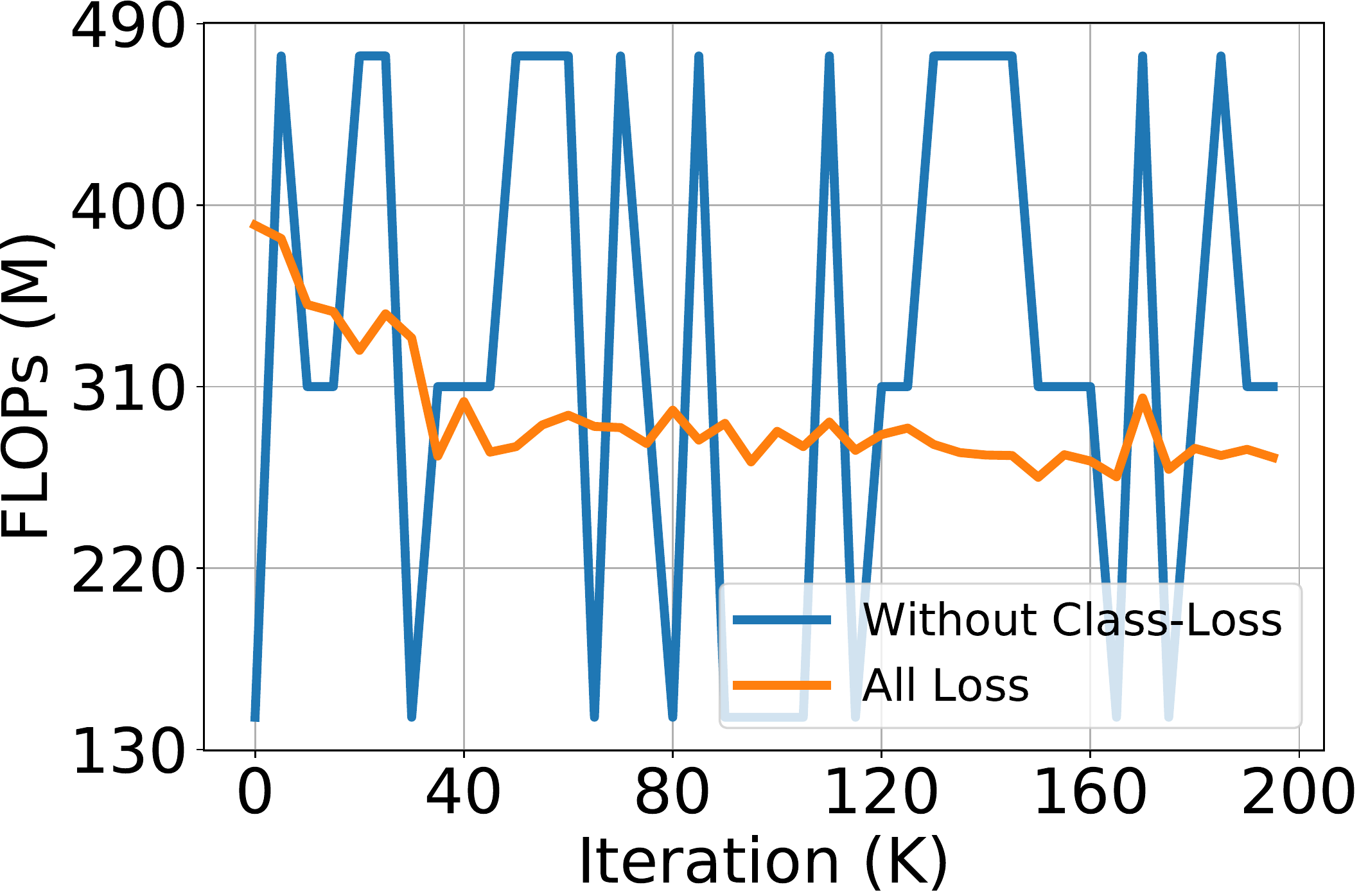}}
  \vskip -0.1cm
  \caption{Training curves comparison of Class-Module with/without Class-Loss for ClassSR-FSRCNN.}
  \label{fig:subclass}
  \vskip -0.8cm
  \end{figure}

In the ablation study, we test the effect of different components and settings with ClassSR-FSRCNN. First, we test the effect of the proposed Class-Loss by removing it from the loss function ($w_2=0$). Fig.~\ref{fig:subclass} shows the curves of PSNR and FLOPs during training. Without the Class-Loss, both two curves cannot converge. This is because that the output probability vectors of the Class-Module all become $[0.333,0.333,0.333]$ under the influence of the Average-Loss. In other words, the input images are randomly assigned to an SR branch, leading to unstable performance. This demonstrates the importance of Class-Loss.

\subsubsection{Effect of Average-Loss}

\label{sec:Effect of Average-Loss}

\begin{figure}[ht!]
  \centering
  \subfigure[PSNR curves]{
  \label{fig:subaveragea}
  \includegraphics[height=2.5 cm]{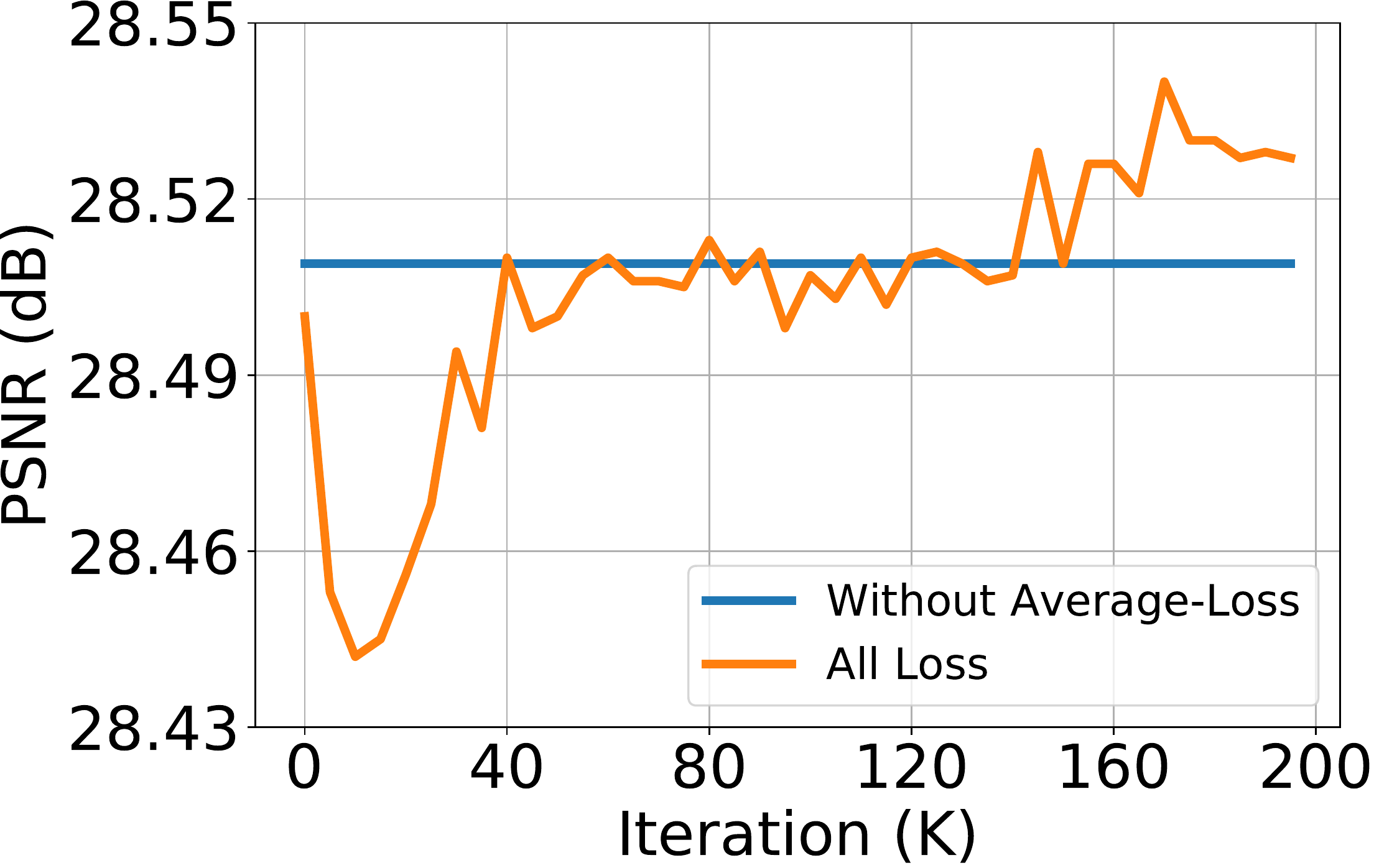}}
  \subfigure[FLOPs curves]{
  \label{fig:subaverageb}
  \includegraphics[height=2.5 cm]{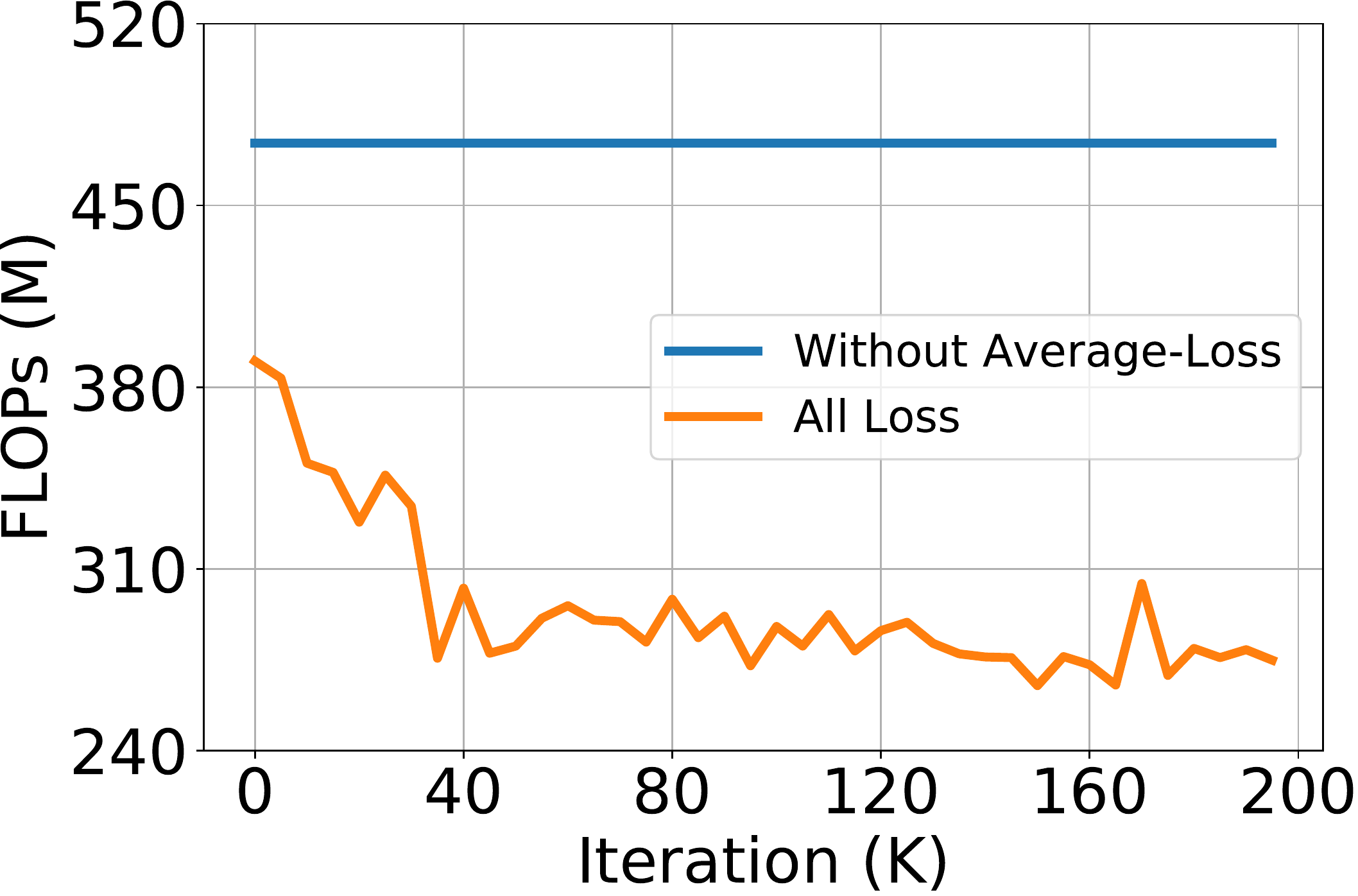}}

  \caption{Training curves comparison of Class-Module with/without Average-Loss for ClassSR-FSRCNN.}
  \label{fig:subaverage}
  \vskip -0.6cm
  \end{figure}

Then we evaluate the effect of the Average-Loss by removing it from the loss function ($w_3=0$). From Fig.~\ref{fig:subaverage}, we can see that both PSNR and FLOPs stop changing from a very early stage. The reason is that all input images are assigned to the most complex branch, which is a bad local minimum for optimization. The Average-Loss is proposed to avoid such biased classification results.

\subsubsection{Effect of the number of classes}

We also investigate the effect of the number of classes, which is also the number of SR branches. We conduct experiments with 2, 3, 4, 5 classes. To pre-train SR branches, we also divide the training data into different numbers of classes, using the same equal-division strategy as in Sec.~\ref{sec:Training Data}. Correspondingly, we set different channels numbers for different settings, as shown in Table~\ref{table:3}. From the results, we can observe that more classes will bring better performance. However, the differences are insignificant. Even the case with two classes achieves satisfactory results. This shows that the ClassSR is robust to the number of classes.

\subsubsection{Controling network complexity in other ways}
\label{sec:layers}

As mentioned in Sec.~\ref{sec:SR-Module}, we obtain branch networks with different network complexities by changing the number of channels and layers at the same time. As shown in Table~\ref{table:5}, we could obtain a comparable performance as reducing channels in Table~\ref{table:2}. The reason why we do not only reduce the layers is that the FLOPs brought by middle layers account for a small proportion of the total FLOPs in light-weight networks (3\% for FSRCNN, 58\% for CARN and 47\% for SRResNet). In other words, even removing all the middle layers can only reduce little FLOPs. Therefore, it is essential to select proper ways to reduce the network complexity for different base networks.

\subsection{ClassSR in other low-level tasks}
\label{sec:ClassSR in other low-level tasks}

To demonstrate that our proposed ClassSR is flexible and can be easily applied to deal with other low-vision tasks, where different regions have different restoration difficulties, we conduct experiments on image denoising. We use DnCNN with different channels (38, 52, 64) as the Denoise-Module to replace SR-Module. Then we train the network on DIV2K\footnote{We use 800 training images (0001-0800) in DIV2K with Gaussian noise ($\sigma$=25) and set patch size as $32 \times 32$.} following the above training settings.

\begin{table}[ht]
  \small %
  \begin{center}

  \begin{tabular}{|c|c|c|}
  \hline
  Model &  Test2K/FLOPs &Test4K/FLOPs  \\
  \hline\hline
  DnCNN-O   & 31.20dB/1.14G(100\%) &32.26dB/ 1.14G(100\%) \\
  DnCNN-C & 31.23dB/0.83G(73\%) &32.28dB/0.76G(67\%) \\
  \hline
  \end{tabular}
  
\end{center}
\vskip -0.3cm
  \caption{PSNR values on Test2K and Test4K. -O: the original network. -C: Denoise with ClassSR framework.}
  \label{table:6}
  \vskip -0.3cm
  \end{table}

As shown in Table~\ref{table:6}, we evaluate the network on Test2K and Test4K with the same noise level. DnCNN with ClassSR framework can obtain higher PSNR than the original DnCNN but with lower computational cost. Compared with SR tasks, there are no enough ``simple'' sub-images in a noisy image in denoising. Therefore, the computational cost saved by ClassSR is not as much as that in SR task. Nevertheless, this result has illustrated ClassSR can be adapted to other low-level vision tasks.

\section{Conclusion}

In this work, we propose ClassSR with a new classification method and two novel loss functions, which could accelerate almost all learning-based SR methods on large images (2K-8K). The key idea is using a Class-Module to classify the sub-images into different classes (e.g., ``simple, medium, hard''), each class corresponds to different processing branches with different network capacity. Extensive experiments well demonstrate that ClassSR can accelerate most existing methods on different datasets. Processing images with more ``simple'' regions (e.g, DIV8K) will save more FLOPs. Besides, ClassSR can also be applied in other low-level vision tasks. 

{\bf Acknowledgements.} This work was supported in part by the Shanghai Committee of Science and Technology, China (Grant No. 20DZ1100800), in part by the National Natural Science Foundation of China under Grant (61906184), Science and Technology Service Network Initiative of Chinese Academy of Sciences (KFJ\-STS\-QYZX\-092), Shenzhen Institute of Artificial Intelligence and Robotics for Society.

{\small
\bibliographystyle{ieee_fullname}
\bibliography{egbib}

\begin{thebibliography}{10}\itemsep=-1pt

\bibitem{DIV2K}
Eirikur Agustsson and Radu Timofte.
\newblock Ntire 2017 challenge on single image super-resolution: Dataset and
  study.
\newblock In {\em Proceedings of the IEEE Conference on Computer Vision and
  Pattern Recognition (CVPR) Workshops}, July 2017.

\bibitem{CARN}
Namhyuk Ahn, Byungkon Kang, and Kyung-Ah Sohn.
\newblock Fast, accurate, and lightweight super-resolution with cascading
  residual network.
\newblock In {\em Proceedings of the European Conference on Computer Vision
  (ECCV)}, pages 252--268, 2018.

\bibitem{set5}
Marco Bevilacqua, Aline Roumy, Christine Guillemot, and Marie~Line
  Alberi-Morel.
\newblock Low-complexity single-image super-resolution based on nonnegative
  neighbor embedding.
\newblock 2012.

\bibitem{SAN}
Tao Dai, Jianrui Cai, Yongbing Zhang, Shu-Tao Xia, and Lei Zhang.
\newblock Second-order attention network for single image super-resolution.
\newblock In {\em Proceedings of the IEEE conference on computer vision and
  pattern recognition}, pages 11065--11074, 2019.

\bibitem{SRCNN}
Chao Dong, Chen~Change Loy, Kaiming He, and Xiaoou Tang.
\newblock Image super-resolution using deep convolutional networks.
\newblock {\em IEEE transactions on pattern analysis and machine intelligence},
  38(2):295--307, 2015.

\bibitem{FSRCNN}
Chao Dong, Chen~Change Loy, and Xiaoou Tang.
\newblock Accelerating the super-resolution convolutional neural network.
\newblock In {\em European conference on computer vision}, pages 391--407.
  Springer, 2016.

\bibitem{DIV8K}
S. {Gu}, A. {Lugmayr}, M. {Danelljan}, M. {Fritsche}, J. {Lamour}, and R.
  {Timofte}.
\newblock Div8k: Diverse 8k resolution image dataset.
\newblock In {\em 2019 IEEE/CVF International Conference on Computer Vision
  Workshop (ICCVW)}, pages 3512--3516, 2019.

\bibitem{ResNet}
Kaiming He, Xiangyu Zhang, Shaoqing Ren, and Jian Sun.
\newblock Deep residual learning for image recognition.
\newblock In {\em Proceedings of the IEEE Conference on Computer Vision and
  Pattern Recognition (CVPR)}, June 2016.

\bibitem{IMDN}
Zheng Hui, Xinbo Gao, Yunchu Yang, and Xiumei Wang.
\newblock Lightweight image super-resolution with information
  multi-distillation network.
\newblock In {\em Proceedings of the 27th ACM International Conference on
  Multimedia}, pages 2024--2032, 2019.

\bibitem{VDSR}
Jiwon Kim, Jung Kwon~Lee, and Kyoung Mu~Lee.
\newblock Accurate image super-resolution using very deep convolutional
  networks.
\newblock In {\em Proceedings of the IEEE conference on computer vision and
  pattern recognition}, pages 1646--1654, 2016.

\bibitem{ADAM}
Diederik~P Kingma and Jimmy Ba.
\newblock Adam: A method for stochastic optimization.
\newblock {\em arXiv preprint arXiv:1412.6980}, 2014.

\bibitem{LapSRN}
Wei-Sheng Lai, Jia-Bin Huang, Narendra Ahuja, and Ming-Hsuan Yang.
\newblock Deep laplacian pyramid networks for fast and accurate
  super-resolution.
\newblock In {\em Proceedings of the IEEE conference on computer vision and
  pattern recognition}, pages 624--632, 2017.

\bibitem{SRResNet}
Christian Ledig, Lucas Theis, Ferenc Husz{\'a}r, Jose Caballero, Andrew
  Cunningham, Alejandro Acosta, Andrew Aitken, Alykhan Tejani, Johannes Totz,
  Zehan Wang, et~al.
\newblock Photo-realistic single image super-resolution using a generative
  adversarial network.
\newblock In {\em Proceedings of the IEEE conference on computer vision and
  pattern recognition}, pages 4681--4690, 2017.

\bibitem{EDSR}
Bee Lim, Sanghyun Son, Heewon Kim, Seungjun Nah, and Kyoung~Mu Lee.
\newblock Enhanced deep residual networks for single image super-resolution.
\newblock In {\em The IEEE Conference on Computer Vision and Pattern
  Recognition (CVPR) Workshops}, July 2017.

\bibitem{RFA}
Jie Liu, Wenjie Zhang, Yuting Tang, Jie Tang, and Gangshan Wu.
\newblock Residual feature aggregation network for image super-resolution.
\newblock In {\em Proceedings of the IEEE/CVF Conference on Computer Vision and
  Pattern Recognition (CVPR)}, June 2020.

\bibitem{pytorch}
Adam Paszke, Sam Gross, Soumith Chintala, Gregory Chanan, Edward Yang, Zachary
  DeVito, Zeming Lin, Alban Desmaison, Luca Antiga, and Adam Lerer.
\newblock Automatic differentiation in pytorch.
\newblock 2017.

\bibitem{RAISR}
Yaniv Romano, John Isidoro, and Peyman Milanfar.
\newblock Raisr: rapid and accurate image super resolution.
\newblock {\em IEEE Transactions on Computational Imaging}, 3(1):110--125,
  2016.

\bibitem{ESPCN}
Wenzhe Shi, Jose Caballero, Ferenc Husz{\'a}r, Johannes Totz, Andrew~P Aitken,
  Rob Bishop, Daniel Rueckert, and Zehan Wang.
\newblock Real-time single image and video super-resolution using an efficient
  sub-pixel convolutional neural network.
\newblock In {\em Proceedings of the IEEE conference on computer vision and
  pattern recognition}, pages 1874--1883, 2016.

\bibitem{SFTGAN}
Xintao Wang, Ke Yu, Chao Dong, and Chen~Change Loy.
\newblock Recovering realistic texture in image super-resolution by deep
  spatial feature transform.
\newblock In {\em Proceedings of the IEEE Conference on Computer Vision and
  Pattern Recognition (CVPR)}, June 2018.

\bibitem{ESRGAN}
Xintao Wang, Ke Yu, Shixiang Wu, Jinjin Gu, Yihao Liu, Chao Dong, Yu Qiao, and
  Chen Change~Loy.
\newblock Esrgan: Enhanced super-resolution generative adversarial networks.
\newblock In {\em Proceedings of the European Conference on Computer Vision
  (ECCV)}, pages 0--0, 2018.

\bibitem{L1}
Zhou Wang, Alan~C Bovik, Hamid~R Sheikh, and Eero~P Simoncelli.
\newblock Image quality assessment: from error visibility to structural
  similarity.
\newblock {\em IEEE transactions on image processing}, 13(4):600--612, 2004.

\bibitem{set14}
Jianchao Yang, John Wright, Thomas~S Huang, and Yi Ma.
\newblock Image super-resolution via sparse representation.
\newblock {\em IEEE transactions on image processing}, 19(11):2861--2873, 2010.

\bibitem{Yu_2018_CVPR}
Ke Yu, Chao Dong, Liang Lin, and Chen~Change Loy.
\newblock Crafting a toolchain for image restoration by deep reinforcement
  learning.
\newblock In {\em Proceedings of the IEEE Conference on Computer Vision and
  Pattern Recognition (CVPR)}, June 2018.

\bibitem{Path-restore}
Ke Yu, Xintao Wang, Chao Dong, Xiaoou Tang, and Chen~Change Loy.
\newblock Path-restore: Learning network path selection for image restoration.
\newblock {\em arXiv preprint arXiv:1904.10343}, 2019.

\bibitem{RCAN}
Yulun Zhang, Kunpeng Li, Kai Li, Lichen Wang, Bineng Zhong, and Yun Fu.
\newblock Image super-resolution using very deep residual channel attention
  networks.
\newblock In {\em Proceedings of the European Conference on Computer Vision
  (ECCV)}, pages 286--301, 2018.

\bibitem{RDN}
Yulun Zhang, Yapeng Tian, Yu Kong, Bineng Zhong, and Yun Fu.
\newblock Residual dense network for image super-resolution.
\newblock In {\em Proceedings of the IEEE conference on computer vision and
  pattern recognition}, pages 2472--2481, 2018.

\bibitem{PAN}
Hengyuan Zhao, Xiangtao Kong, Jingwen He, Yu Qiao, and Chao Dong.
\newblock Efficient image super-resolution using pixel attention, 2020.

\end{thebibliography}
}

\newpage
\cleardoublepage

\includepdf[pages={1}]{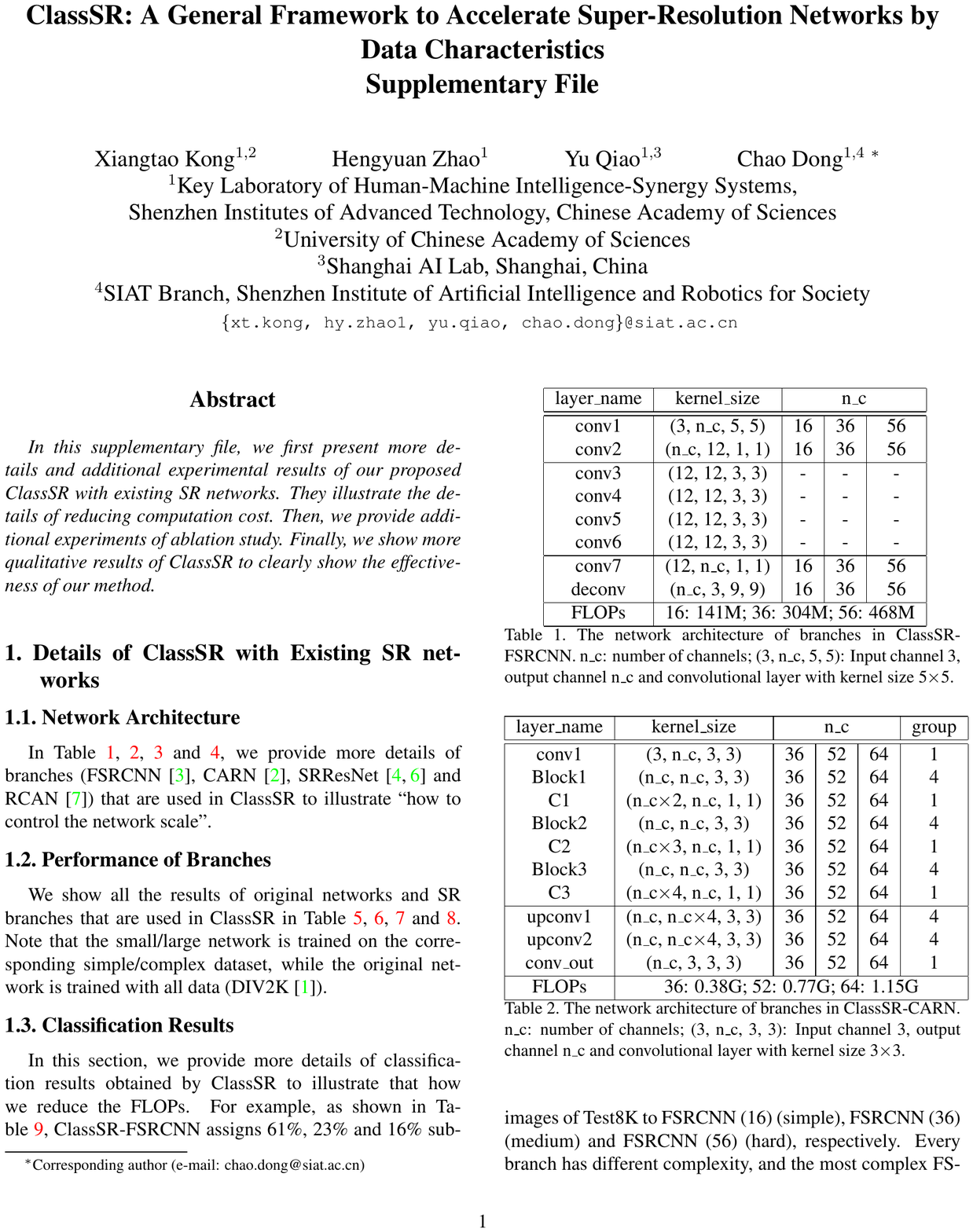}
\includepdf[pages={2}]{sup_ClassSR-CVPR-arxiv.pdf}
\includepdf[pages={3}]{sup_ClassSR-CVPR-arxiv.pdf}
\includepdf[pages={4}]{sup_ClassSR-CVPR-arxiv.pdf}
\includepdf[pages={5}]{sup_ClassSR-CVPR-arxiv.pdf}
\includepdf[pages={6}]{sup_ClassSR-CVPR-arxiv.pdf}

\end{document}